# RAVQ-HoloNet: Rate-Adaptive Vector-Quantized Hologram Compression

Shima Rafiei[1], Zahra Nabizadeh Shahr-Babak[1], Soroush Khoubyarian[2], Alexandre Cooper[3], Shadrokh Samavi[4], and Shahram Shirani[1]

[1]Department of Electrical and Computer Engineering, McMaster University, Hamilton, Ontario, Canada
[2]Department of Physics and Astronomy, University of Waterloo, Waterloo, Ontario, Canada
[3]Institute for Quantum Computing, Department of Physics and Astronomy, University of Waterloo, Waterloo, Ontario, Canada
[4]Computer Science Department, Seattle University, Seattle, Washington, USA

**Abstract**

Holography offers significant potential for AR/VR applications. However, its adoption is limited by the high demand for data compression. Existing deep learning approaches generally lack rate adaptivity within a single network and often require multiple models to cover different bandwidth requirements. We present RAVQ-HoloNet, a rate-adaptive vector quantization framework that integrates the rate-adaptive compression with the transformation of image data into phase-only hologram. RAVQ-HoloNet achieves high-fidelity reconstructions, outperforming current state-of-the-art methods implemented via two distinct architectural configurations: a standard model optimized for low bit rates and a deeper, extended variant tailored for ultra low bit rate setting. To evaluate these models, we utilized the DIV2K dataset as a benchmark for high-fidelity holographic reconstruction. Quantitative analysis in the simulation reveals that our approach significantly surpasses current benchmarks. Specifically, in the low bit rate domain, our method achieves a BD-Rate reduction of −33.91% and a BD-PSNR gain of 1.02 dB relative to the state-of-the-art method. Additionally, experimental results on the SLM device show that our method achieves higher contrast and improved quality.

# 1 Introduction

With the growing demand for immersive experiences in applications such as metaverse, Augmented Reality (AR) and Virtual Reality (VR), holography has emerged as a key technology for generating realistic three-dimensional (3D) content. Holographic displays generate a 3D scene by diffracting light through a hologram plane. The hologram plane serves as a medium that encodes the interference patterns and captures the light field corresponding to a 3D scene. When illuminated, the hologram reconstructs the original wavefronts of the light. As these wavefronts reach the observer's eyes, they are perceived as if a real 3D object, creating the illusion of a 3D scene [1][2]. The hologram, which preserves both the amplitude and phase information of the light field, can be created through optical setups or numerical simulations, known as Computer-Generated Holography (CGH). CGH enables the synthesis of holograms through numerical simulations or deep learning techniques, eliminating the need for complex optical setups. During the past decades, significant efforts have been devoted to developing methods for CGH.

## 1.1 Hologram Computation

Among the computational approaches to CGH, traditional methods such as point-source and wavefront propagation represent the object scene based on its wave-emitting properties. Based on the Huygens-Fresnel principle, the point source method considers an object as a dense array of wave emitters, each radiating a spherical wavefront to construct the hologram via interference [1]. This approach which is typically suited for point-cloud objects, offers a high reconstruction fidelity at the cost of significant computational complexity. Alternatively, layer-based methods use coherent planar wavefronts to model uniform surfaces efficiently, typically using Fresnel diffraction for near-field and the Angular Spectrum Method (ASM) for far-field scenarios [3]. This approach, typically suitable for RGB-D and multilayer imaging. These methods commonly apply Fast Fourier Transforms (FFTs) to efficiently com-pute the wave propagation via a propagator function in the frequency domain. Polygon-based approaches [4] further bridge the gap between complexity and efficiency by combining FFTs with coordinate transformations to handle diffraction from tilted surfaces. Despite their effectiveness, object-based representation methods are computationally demanding for large-scale dynamic scenes. Recently, learning-based approaches have gained significant attention in CGH. For example, [5] proposed a convolutional neural network (CNN) with focal stack loss that generates photorealistic, complex-valued 3D holograms directly in real time from polygonal mesh models reconstructed from captured point clouds.

In practical holographic display systems, Spatial Light Modulators (SLMs) are employed to modulate the wavefront of light for 3D scene reconstruction. However, most commercially available SLMs are limited to phase-only modulation because of their high diffraction efficiency and hardware simplicity [6].

When complex-valued object waves are available, methods such as double-phase techniques [7, 8] or deep learning-based approaches [9, 10] are used to directly convert the complex field into a phase-only hologram. For instance, Double-Phase Amplitude Coding (DPAC) achieves this by mathematically decomposing a complex wavefront into two phase-only components with a constant amplitude, thereby facilitating accurate reconstruction without iterative computation. [5] proposed an Anti-Aliasing Double-Phase method (AA-DPM), a modified DPAC approach designed specifically for CNN-predicted complex holograms, which effectively mitigates aliasing artifacts around high-frequency details and occlusion boundaries. Other methods, such as direct deep learning, aim to predict phase-only holograms, typically in an unsupervised manner. These approaches typically employ an encoder–decoder architecture trained to map intensity distributions into phase-only holograms. Recent unsupervised methods include the HoloNet [9], DeepCGH [10], DGH [11], and 3D-DGH [12]. The method proposed in [6] employs supervised learning followed by unsupervised learning. In unsupervised settings [9, 10], CNNs are trained to generate phase-only holograms using only the target intensity (amplitude) in one or more depth planes for supervision. Without access to the ground-truth phase, the network autonomously learns to produce an appropriate phase-only hologram that can reconstruct a desired intensity. It is analogous to traditional optimization-based approaches such as Stochastic Gradient Descent (SGD) or the Gerchberg–Saxton (GS) method as phase retrieval algorithms [13], where the complex object wave is unknown, and a randomly initialized phase is iteratively updated—either through gradient backpropagation in the case of SGD-based methods or through alternating projections in the GS algorithm—to reconstruct a phase pattern that yields the desired target intensity upon propagation. However, both iterative algorithms based on the GS method and SGD-based approaches, such as WirtingerSGD [14], are computationally intensive and require numerous iterations to achieve convergence.

## 1.2 Hologram Compression

To enable real-time holographic applications, hologram compression is essential for reducing storage, transmission, and processing costs [15], along with optimized computational pipelines, particularly for AR/VR near-eye holographic displays [16]. [8] demonstrated the feasibility of AR/VR systems by shifting the optical complexity to software through Fresnel holography to synthesize object wavefronts, thereby allowing improved control over most system constraints.

However, as holography advances toward practical deployment in AR/VR, it paves the way for real-time applications and interactive entertainment, where low latency and high visual fidelity are essential [16].

Despite the growing interest in holographic displays, only a few research methods have focused on hologram compression. Dual Phase Retrieval and Compression (DPRC) [16] introduced a unified framework that integrates hologram generation and compression, in which the computationally intensive hologram generation is offloaded to a sender, such as a cloud server (encoder). Lightweight decoding was performed on a receiver edge device. This method enables a fast and efficient data transfer by transmitting a compact latent representation. Other methods typically treat hologram computation and compression separately. A common approach is to compute the hologram and then apply standard image or video codecs for compression. However, traditional codecs, such as JPEG, HEVC, and H.264, are ill-suited for holographic content. High-frequency interference patterns and phase-dependent details intrinsic to holograms often lead to significant quality degradation when compressed using general-purpose standards [17]. Therefore, learning-based methods have demonstrated high success rates. [18] introduces a deep learning framework for phase-only holograms by employing a phase-specific distortion metric and entropy bottleneck. Their method outperforms traditional codecs such as VVC in rate-distortion performance on the non-public

ETRI dataset. [19] also proposes a hybrid approach for holographic complex-valued video compression, combining H.265 video coding with the HiFiHC neural network as a conditional Generative Adversarial Network (GAN). Motion vectors are used to predict inter-frame content, whereas a deep neural network encodes residual errors to capture fine holographic details, enabling the efficient compression of temporally coherent holographic sequences. The latent space of the network is compressed using the Ballé method. [20] divides the holographic content into distinct foveated and peripheral areas as input into two separate deep learning pathways. It applies higher compression to peripheral areas, where human vision is less sensitive, while preserving high quality in foveated regions. Although recent neural compression methods for holograms have shown promising results, most are fundamentally based on the Ballé-style framework, which utilizes continuous latent variables, hyperprior models, and entropy bottlenecks for compression [18, 16, 19, 20]. These approaches follow the rate-distortion optimization principles laid out in [21, 22], adapting them to holographic data.

In this paper, we propose a pipeline for joint hologram computation and compression in an unsupervised manner similar to HoloNet [9], aiming to generate phase-only holograms with built-in compression inspired by [23]. We proposed a VAE-inspired architecture for end-to-end optimized hologram compression. The main contributions of this study are as follows:

- We propose unsupervised hologram compression based on a hierarchical VQ-HoloNet architecture, which learns discrete latent representations without ground-truth phase supervision.
- The model is rate-adaptive through its multi-level vector quantization structure and adaptive codebook size, enabling control over the compression rate and reconstruction quality.
- Our hologram compressor accepts either complex holograms or intensity-only inputs and compresses them, subsequently decompressing data into phase-only holograms. Therefore, the proposed method can operate as either a hologram generator–compressor or a standalone compressor, depending on the type of input provided.
- Our proposed method is further demonstrated through physical SLM implementation.

The remainder of this paper is structured as follows: Section II presents the proposed method, including its architecture and the loss functions used. Section III explains the compression viewpoint, while Section IV describes the training and loss function. Section V is the implementation configuration and analyzes of the simulation results, whereas section VI details the hardware setup and experimental results, followed by Section VII, which explores future work. Finally, the last section summarizes the conclusions of this study.

# 2 Proposed Method

Holographic data inherently require significant storage and transmission bandwidth. Therefore, efficient compression is critical for enabling holographic applications, especially for resource-constrained edge devices. Our proposed unsupervised framework integrates two key components: (1) an initial complex hologram generation stage based on the U-Net. This stage can be bypassed if a complex hologram is available rather than the object intensity, and (2) hologram compression via a Rate Adaptive Vector Quantized HoloNet (RAVQ-HoloNet), where a complex-valued encoder and phase-only hologram decoder are unified within a single network. As shown in Fig. 1, our RAVQ-HoloNet model consists of a VQ-HoloNet, depicted in the middle box, and two sequence-to-sequence (Seq2Seq) models at the top and bottom levels.

Inspired by the VQ-VAE-2 architecture [24], VQ-HoloNet is composed of three main components: a complex-valued encoder, vector quantization module, and lightweight decoder that generates the final phase-only holograms. In the VQ-HoloNet part, the encoder operates directly on the input complex-valued hologram, and extracts features from this complex input. The decoder processes these features to generate a phase-only hologram. The encoding and decoding process is done in two hierarchical levels, top and bottom. Hierarchical latent representations are widely used in state-of-the-art image compression frameworks for natural images [24], as they enable the model to simultaneously capture global structures and local details, leading to high-fidelity reconstructions.

In conventional VQ-VAE models, the vector quantization component lacks flexibility in adjusting the compression levels and typically requires retraining for each desired bitrate or compression ratio. This results in the need for separate networks for each compression setting, which requires extra storage to maintain multiple encoders on the sender side and corresponding decoders on the receiver side. This is not suited for storage-restricted edge devices. To overcome this limitation, we incorporate Seq2Seq models, as originally introduced in [23], to compress the codebooks produced by our VQ-HoloNet at multiple bitrates. This enables rate-adaptive compression and completes the RAVQ-HoloNet architecture. A block diagram of the proposed framework is shown in Fig. 1. Details of components of each part, hologram generation and hologram compression are described in the following sections.

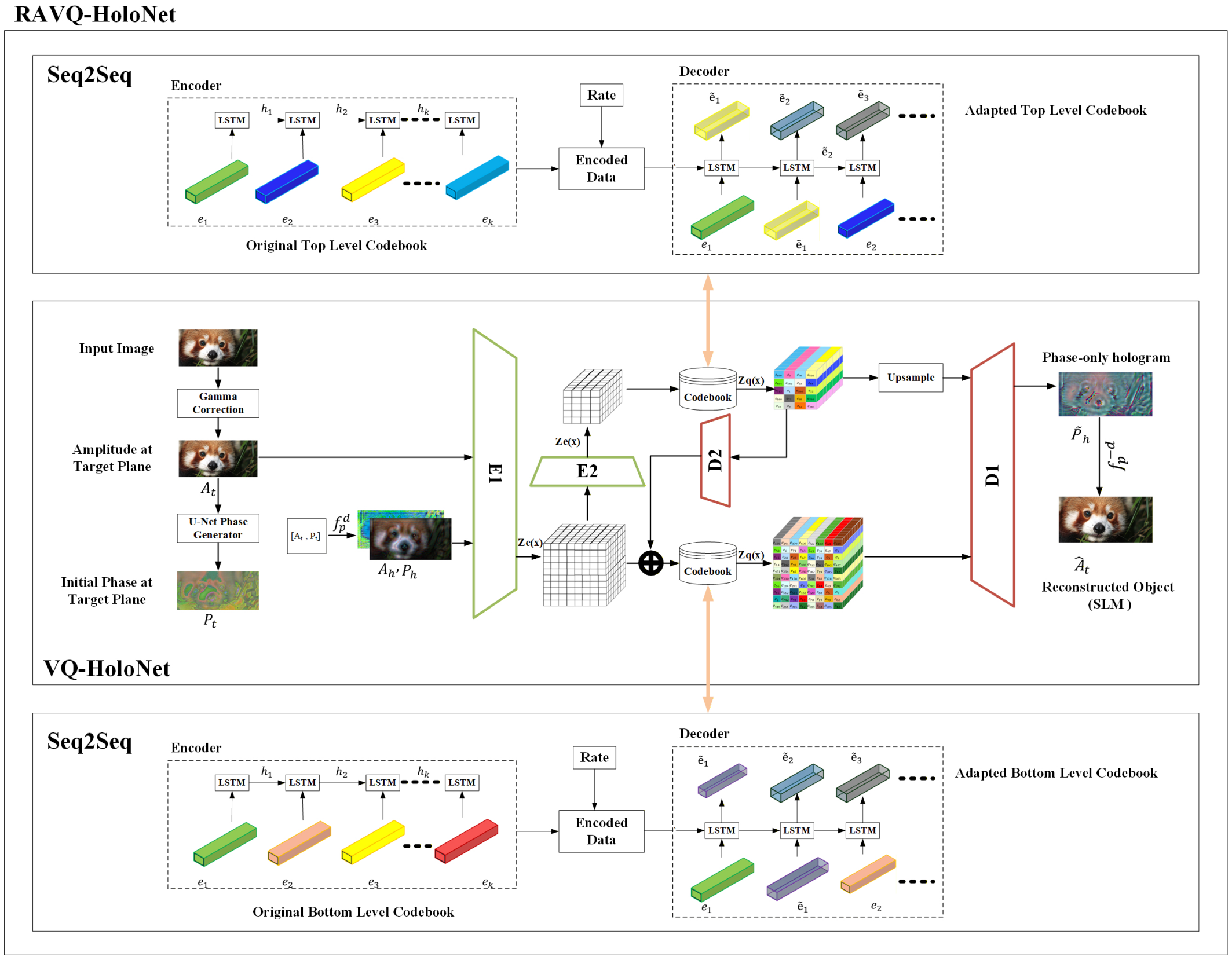

Figure 1: The block diagram of the proposed method. Our framework, RAVQ-HoloNet, incorporates a central VQ-HoloNet and two Seq2Seq models. The VQ-HoloNet generates a phase-only hologram from an input intensity image through a multi-stage process. First, a U-Net generates an initial phase $P_t$, followed by free-space light propagation $f_p^d$ to the complex-valued hologram plane $(A_h,\ P_h)$. Finally, the network encodes the resulting complex-valued wavefront into a quantized latent space (code vectors) and decode it into phase-only hologram $\tilde{P}_h$. Furthermore, each Seq2Seq encoder receives the original codebook used by the latent space $\mathbf{e} = \{e_1, e_2, \ldots, e_K\}$ and produces an intermediate representation (Encoded Vector). The subsequent Seq2Seq decoder transforms this representation into the adapted codebook $\tilde{\mathbf{e}} = \{\tilde{e}_1, \tilde{e}_2, \ldots, \tilde{e}_{\tilde{K}}\}$, enabling a reduction in codebook size when $\tilde{K} < K$.

## 2.1 Complex Hologram Generation

In the first stage of the proposed framework, an intensity image is used as the input. This image is initially transformed into a target amplitude ($A_t$) by applying the inverse of gamma correction, followed by a square-root operation, as described in [9]. An inverse gamma correction is used to counteract the gamma adjustment performed by the camera. The resulting amplitude is then used to generate an initial phase estimate ($P_t$) using a pre-trained U-Net from [16]. These two components, the target amplitude and generated phase, are combined to form a complex-valued light field in the object domain (target plane). This complex field is subsequently propagated to the hologram plane using a bandwidth-limited ASM with propagation distance of $d$. The propagated field provides both the amplitude ($A_h$) and phase ($P_h$) at the hologram plane.

Because ASM inherently has information loss when propagating the complex-valued field from the object to the hologram plane, the reconstruction accuracy of the object can be limited [3]. We mitigate this by providing VQ-HoloNet with dual-domain inputs. Specifically, the network receives the amplitude of the object domain alongside the complex-valued hologram from the hologram plane. This approach enables the network to decode a more accurate phase-only output, resulting in an improved reconstruction of the desired intensity in the object domain.

The mathematical formulation of the ASM, which is used to simulate the wave propagation, is presented in the loss function section, because it is essential for computing the reconstruction loss in the object domain.

## 2.2 Hologram Compression

In the compression stage of our framework, the hologram is encoded into a compact representation referred to as the latent space, which contains all the information required for subsequent reconstruction. To achieve this, we propose an encoder-decoder-based architecture, VQ-HoloNet, which compresses complex-valued holographic data and reconstructs phase-only output. VQ-HoloNet employs two hierarchically multi-scale encoders, $E_1$ (bottom encoder) and $E_2$ (top encoder), together with their respective decoders, $D_1$ (bottom decoder) and $D_2$ (top decoder). This hierarchical structure operates as follows (see Fig. 1): The top-level latent representation (E2 output) is first quantized and decoded by D2. This decoded representation is then concatenated with the bottom-level latent (E1 output) to become jointly quantized. After the combined latent is quantized, it merges with the quantized upsampled top latent for joint decoding (D2). This hierarchical process allows the model to capture both high-level and detailed hologram information. Figs. 2 illustrates the architecture of the E1 and E2 encoders. The encoder consists of convolutions and four residual blocks [25], which can mitigate the instability of the gradient for deep-layered structures. Encoder outputs are quantized based on codebooks. In Fig. 3, D1 and D2 consist of the same number of residual blocks, followed by transposed convolution layers.

To make the latent space suitable for efficient storage and transmission, it is necessary to discretize the continuous latent space. For this purpose, we employ vector quantization. Vector quantization is particularly well-suited to holographic data, as previous studies [26] have shown that holograms exhibit low histogram variation compared to natural images. This property results in statistical redundancy in the complex wavefronts, making vector quantization an effective choice for compression.

To perform quantization, we utilize the Exponential Moving Average (EMA) clustering algorithm [27]. This algorithm assigns each latent vector to the nearest entry in a learnable codebook, where each codebook entry (code vector) acts as the cluster centroid. During training, the codebook is constantly updated to represent the most significant features in the data.

# 3 Compression Viewpoint

The hologram is compressed at the sender and decompressed at the receiver, which may be an edge client [16]. Fig. 4 illustrates this data transfer process. Within our framework, the compression ratio is primarily determined by two factors, the codebook size and the latent space resolution. These components are discussed in detail in the following sections.

To achieve different compression rates for the discrete latent codebooks and enable adaptive rate control, we integrate the VQ-HoloNet structure with an embedded Seq2Seq encoder-decoder [23]. This addition allows the codebooks generated by VQ-HoloNet to be further compressed at multiple bitrates, making the overall framework both flexible and efficient. This integration serves as a bitrate tuner, allowing the framework to dynamically compress discrete latent codebooks by transforming the entries from size $K$ to a reduced size $\tilde{K}$ i.e. $(\tilde{K} < K)$ . By adjusting the number of codebook entries within a single network, our framework achieves a flexible compression range and efficient rate-distortion performance across various bit-rate scenarios.

Unlike Ballé-based compression networks [16], which require the training and storage of separate model weights for each compression level, our method supports a continuous spectrum of bitrate-quality trade-offs using the same model. This flexibility makes our approach particularly suitable for adaptive compression scenarios, such as real-time quality-of-service (QoS) management over variable network conditions. The adapted codebook, denoted as $\tilde{\mathbf{e}} = \{\tilde{e}_1, \tilde{e}_2, \ldots, \tilde{e}_{\tilde{K}}\}$, is generated by the Seq2Seq model, which utilizes a series of Long Short-Term Memory (LSTM) [28, 29]. LSTM units encode original codebook $\mathbf{e} = \{e_1, e_2, \ldots, e_K\}$ into a latent encoded vector and decode it sequentially to new code vectors. In this context, code vectors are treated analogous to words in a sentence, allowing the LSTM encoder to summarize the original codebook and the LSTM decoder to reconstruct a compact codebook of the desired size. The target codebook size, $\tilde{K}$, can be specified by the user or dynamically selected based on network requirements.

During training, our RAVQ-HoloNet is optimized using various codebook sizes to ensure robustness and adaptability. The Seq2Seq model applies the teacher-forcing strategy [30], in which ground-truth information is provided to the decoder in odd steps of decoding prediction for half the size of $\tilde{K}$. In even steps and the following half size, the decoder relies on the input code vectors. Periodically providing ground-truth data as a reference improves the stability and convergence of the training. At inference time, the Seq2Seq model dynamically generates the adapted codebook $\tilde{\mathbf{e}}$, thus matching the output bitrate to the current demand.

The codebook size selection has to be based on QoS. Typically, higher QoS employs larger codebook sizes. As illustrated in Fig. 4, codebooks of various sizes are pre-generated and stored at both the sender and receiver, eliminating the need to regenerate them using the Seq2Seq model for each transmission. In other words, Seq2Seq is utilized offline to derive the codebooks. This strategy minimizes computational latency and reduces transmission time. However, in case of any change in prestored codebooks, there is no need to retrain the model. Our decoder is agnostic to the codebook size which can be any value. The system only requires that the updated codebook be synchronized across both the encoder and decoder.

Because the necessary codebooks are already stored on the receiver side, only the sequence of corresponding indices must be transmitted, instead of the fully quantized feature vectors [27]. This is highly advantageous, as quantized feature vectors are high-dimensional (e.g., 128 dimensions) and stored with floating-point precision, resulting in much greater transmission costs than the integer indices.

Our current configuration is referred to as the VQ-HoloNet (low) model; its latent space volume is reported in the first column of Table 1 before and after quantization. Given the input size of $H \times W$, the latent representations have spatial latent space resolutions of $H/4 \times W/4$ for the bottom and $H/8 \times W/8$ for the top level. In addition to our VQ-HoloNet (low), we introduce our VQ-HoloNet (ultra low) which has a reduced resolution of the latent space. This configuration is implemented by having a deeper network compared to the low model. By lowering the dimensionality of the latent space, a lower bitrate can be achieved. For VQ-HoloNet (ultra low), the architecture is similar, except that we have one more convolution layer in the encoder before residual blocks and one more transposed convolution layer in the decoder after the residual blocks. This results in half the latent space resolution, $H/8 \times W/8$ for the bottom level and $H/16 \times W/16$ for the top level (second column of Table 1).

The rows in Table 1, report the volume of the latent space when it is first continuous (float32) with 128 channels, after quantization and conversion to indices, the size becomes an integer with index sizes of $log_2\tilde{K}$ where $\tilde{K}$ is the size of codebook. To further enhance the compression efficiency, the indices of code vectors are entropy-coded using the lossless entropy coding of

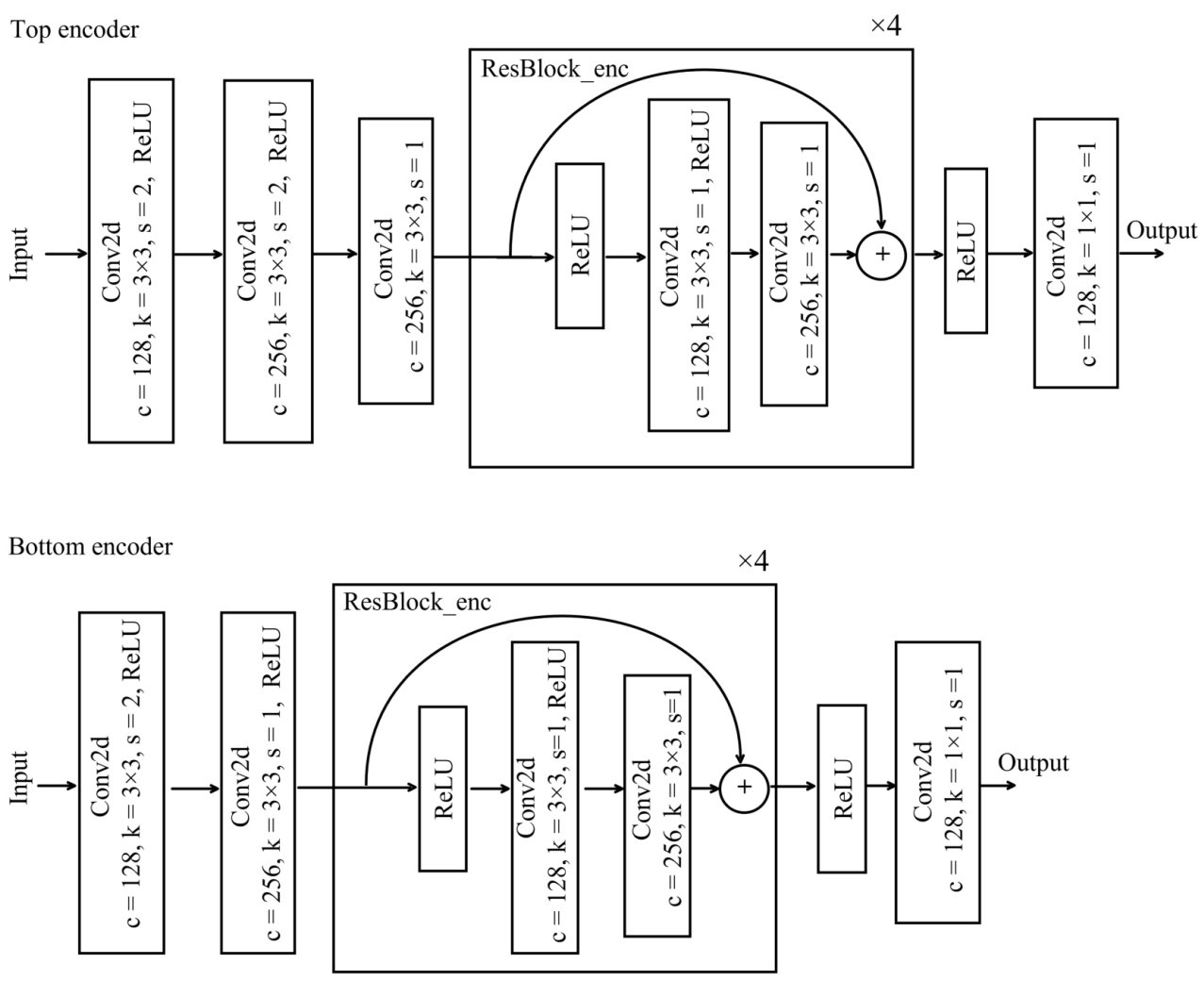

Figure 2: The bottom (E1) and top (E2) encoders in VQ-HoloNet. Parameters $c, k$, and $s$ represent the output channels, kernel size, and stride, respectively. Only $s > 1$ are shown.

Table 1: Latent space resolution: Quantitative comparison of continuous and quantized latent space before entropy coding.

| **Data type** | VQ-HoloNet(low) | VQ-HoloNet(ultra low) |
|---|---|---|
| Continuous (float32) | $32 \times 128 \times \left(\frac{H}{4}\frac{W}{4} + \frac{H}{8}\frac{W}{8}\right)$ | $32 \times 128 \times \left(\frac{H}{8}\frac{W}{8} + \frac{H}{16}\frac{W}{16}\right)$ |
| Quantized (Integer) | $\log_2(\tilde{K}) \times \left(\frac{H}{4}\frac{W}{4} + \frac{H}{8}\frac{W}{8}\right)$ | $\log_2(\tilde{K}) \times \left(\frac{H}{8}\frac{W}{8} + \frac{H}{16}\frac{W}{16}\right)$ |

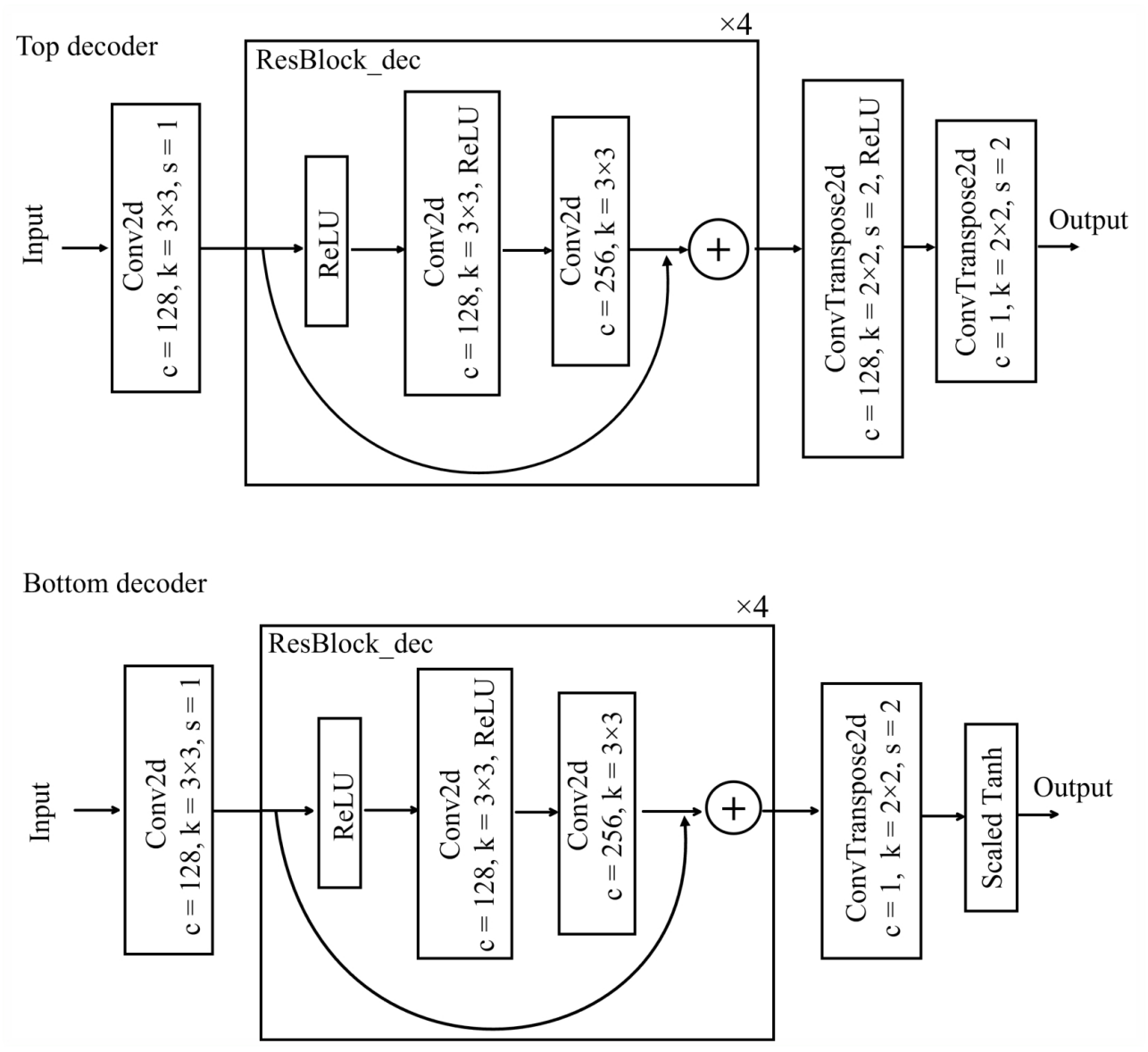

Figure 3: Bottom (D1) and top (D2) decoders in VQ-HoloNet. Parameters $c, k$, and $s$ represent the output channels, kernel size, and stride, respectively. Only $s > 1$ are shown. The final layer of D2 utilizes a scaled tanh activation function with scale factor of $\pi$ to map the reconstructed phase values to the $[-\pi, \pi]$ range for SLM compatibility.

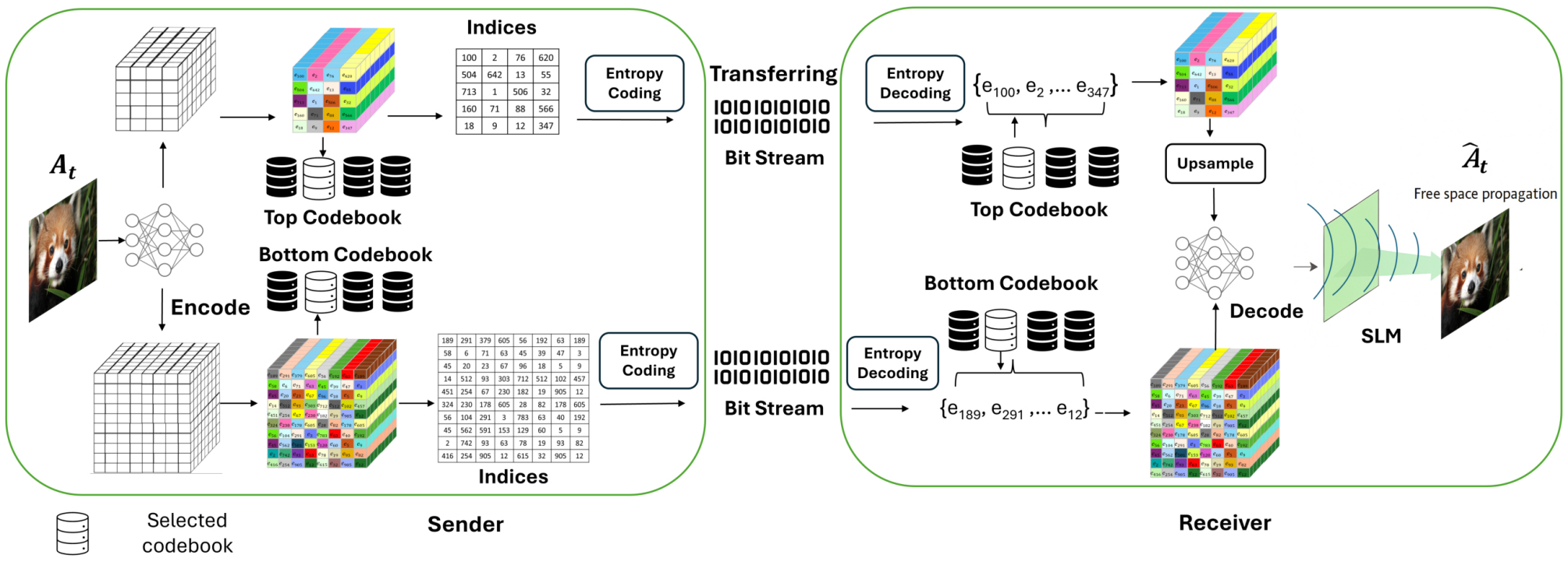

Figure 4: A simplified illustration of the sender-receiver hologram communication framework. Codebooks of different sizes are generated using a Seq2Seq model and stored at both the sender and receiver. The codebook size is selected based on QoS requirements, enabling the receiver to transmit indices corresponding to the chosen codebook.

Asymmetric Numeral Systems (ANS) [31]. ANS is adopted in this framework as it achieves a compression ratio near the theoretical Shannon entropy while maintaining the low computational complexity and high throughput characteristic of Huffman coding. In the proposed pipeline, the indices are transformed into an entropy-coded bitstream for transmission. Upon reception, the system performs lossless entropy decoding to recover the indices, which subsequently serve as pointers to retrieve the corresponding code vectors from the selected codebook.

# 4 Training and Loss Functions

Our model is trained in two stages. During training, the VQ-HoloNet backbone is first trained for $N$ epochs as warm-up training. After this warm-up phase, the Seq2Seq component is activated.
In the first stage, we use the following loss function.

$$\mathcal{L}_{\text{framework}} = \text{Latent}_{\text{VQ}}(\mathbf{e}) + \mathcal{L}_{\text{VQ}_{\text{Reconstruct}}}(\mathbf{e}) \tag{1}$$

The complex hologram $\mathbf{x} = (A_h, P_h)$ is encoded into a latent representation by the encoder and then quantized using a learned codebook; that maps each encoder output to its nearest codebook vector.

$$z_q(\mathbf{x}) = Q(E_\theta(\mathbf{x})|\mathbf{e}) = \arg\min_{0 \le i \le K} \|E_\theta(\mathbf{x}) - e_i\|_2 \tag{2}$$

where $z_q(\mathbf{x})$ is the quantized latent, $E_\theta$ is the encoder, and $e_i$ denotes the $i_{th}$ code vector in the codebook. The decoder $D_\phi$ reconstructs the phase-only hologram $\tilde{P}_h$ from the quantized latent representation:

$$\tilde{P}_h = D_\phi(z_q(\mathbf{x})) \tag{3}$$

Because the codebook is learned during training, we add a dedicated loss to regulate it. Since quantization is non-differentiable, we adopt the straight-through estimator to back-propagating gradients [24]. Therefore, the overall objective includes two components: codebook and a commitment losses.

$$\begin{aligned} \text{Latent}_{\text{VQ}}(\mathbf{e}) = &\underbrace{\|\text{sg}[E_\theta(\mathbf{x})] - z_q(\mathbf{x})\|_2^2}_{\text{Codebook Loss}} \\ &+ \beta \underbrace{\|\text{sg}[z_q(\mathbf{x})] - E_\theta(\mathbf{x})\|_2^2}_{\text{Commitment Loss}} \end{aligned} \tag{4}$$

where sg[·] denotes the stop-gradient operation, and $\beta$ is a hyperparameter.

As shown in Eq. 1, the first stage of training also has reconstruction loss $\mathcal{L}_{\text{VQ}_{\text{Reconstruct}}}(\mathbf{e})$. Due to the high-frequency phase variations in holographic data, neural networks may struggle to learn accurate phase reconstructions. To address this, we incorporate a physics-inspired wave propagation model into our loss function, guiding the network with a differentiable forward model that simulates light propagation. Although the Rayleigh–Sommerfeld (RS) integral [32, 1] is the most accurate, it is computationally expensive. For efficiency, we use the band-limited ASM [3], formulated as:

$$\begin{aligned} f_p^d(\phi, a) = \iint &\mathcal{F}\left[e^{i\phi(x,y)} a(x,y)\right] \\ &\cdot H(f_x, f_y, d) \cdot e^{i2\pi(f_x x + f_y y)}\, df_x\, df_y, \end{aligned} \tag{5}$$

where $\mathcal{F}$ denotes the 2D Fourier transform, $f_x$ and $f_y$ are the spatial frequencies, $a(x, y)$ is the amplitude, and $H(f_x, f_y, d)$ is the kernel function:

$$H(f_x, f_y, d) = \begin{cases} \exp\left( i \, \frac{2\pi d}{\lambda} \, \sqrt{1 - (\lambda f_x)^2 - (\lambda f_y)^2} \right), \\ \quad \text{for } f_x^2 + f_y^2 < \frac{1}{\lambda^2}, \\ 0, \quad \text{otherwise.} \end{cases} \tag{6}$$

$\phi(x, y)$ is a real-valued phase, and $H(f_x, f_y)$ acts as a low-pass filter, removing spatial frequencies beyond the physical cutoff defined by wavelength $\lambda$. The forward model $f_p^d(\phi, a)$ numerically propagates a complex field from the hologram to the target plane. Inverse hologram propagation to the object domain enables us to compare the reconstructed amplitude $|\hat{A}_t| = |f_p^{-d}(\phi, 1)|$ with the ground truth amplitude $A_t$.
For computational efficiency, ASM is expressed in the Fourier domain as follows:

$$f_p^d(\phi, a) = \mathcal{F}^{-1}\left\{\mathcal{F}\left[a(x, y)e^{i\phi(x,y)}\right] \cdot H(f_x, f_y, d)\right\},$$

where $\mathcal{F}^{-1}$ denotes the inverse 2D Fourier transform.
To ensure high-quality phase retrieval, we combine three complementary loss terms between the reconstructed amplitude $|\hat{A}_t| = |f_p^{-d}(\phi, 1)|$ and the target amplitude $A_t$:

- MSE Loss: Pixel-wise mean squared error.
- MS-SSIM: Multi-Scale Structural Similarity Index for perceptual quality [33].
- Watson-DFT: Frequency-domain perceptual loss [34].

The reconstruction loss is:

$$\begin{aligned} \mathcal{L}_{\text{VQ}_{\text{Reconstruct}}}(\mathbf{e}) = {} & a_{\text{MSE}} \cdot \mathcal{L}_{\text{MSE}} + a_{\text{MS-SSIM}} \cdot (1 - \mathcal{L}_{\text{MS-SSIM}}) \\ & + a_{\text{WFFT}} \cdot \mathcal{L}_{\text{WFFT}}, \end{aligned} \tag{7}$$

where $a_{\text{MSE}}, a_{\text{MS-SSIM}}, a_{\text{WFFT}}$ are weighting coefficients. To calculate $\mathcal{L}_{\text{VQ}_{\text{Reconstruct}}}$, the phase-only map is first reconstructed to the object domain using the ASM in Eq. 5, quality metrics are applied between the reconstructed object and original amplitude.
In the second stage of training, Seq2Seq is activated after training in the first stage for N epochs. By activating Seq2Seq, the generated codebook is fed to Seq2Seq, and a new codebook is generated. Thus, the framework loss can be changed as follows.

$$\mathcal{L}_{\text{framework}} = \text{Latent}_{\text{VQ}}(\tilde{\mathbf{e}}) + \mathcal{L}_{\text{VQ}_{\text{Reconstruct}}}(\tilde{\mathbf{e}}) \quad (8)$$

where $\tilde{\mathbf{e}}$ is the output of Seq2Seq. The gradients in the Seq2Seq model are backpropagated through the decoder and latent space toward both the LSTM encoder and decoder, enabling the model to adapt to varying codebook sizes. For the newly introduced codebook sizes, the objective is to achieve a low latent loss while maintaining a high reconstruction quality.

# 5 Simulation Results

This section first provides an overview of the dataset and simulated optical setup used in our experiments. We then describe the network configuration in detail, followed by a comprehensive presentation of both the quantitative and qualitative analyses of our results. The effectiveness of the proposed method is demonstrated and discussed based on various performance metrics and visual comparisons.

## 5.1 Dataset

To fairly evaluate and compare the performance of different compression methods, a public dataset with consistent simulated optical parameters is essential. In our experiments, we utilize the DIV2K dataset [35], which comprises 800 images and is commonly used for compression evaluation. For evaluation, we employ two separate test sets: (i) a 100-image validation subset of DIV2K used by [16], which we denote as Set1, and (ii) the 50-image test subset of DIV2K, denoted as Set2.

In our simulation, the pixel pitch of the hologram plane was set to 6.4 $\mu$m. RGB illumination was employed, using wavelengths of 450 nm (blue), 520 nm (green), and 638 nm (red). The propagation distance between the object plane and the hologram plane was fixed at 20 cm. Owing to the wavelength-dependent nature of the diffraction, a separate model was trained for each color channel.

## 5.2 Development Platform

The proposed method was implemented on a platform equipped with an RTX 3090 GPU (24 GB VRAM) and AMD Ryzen 9 5950X 16-Core Processor CPU.

## 5.3 Comparison with Related Works

Several previous methods have addressed the challenges of hologram compression. [16, 20, 18] follow the approach proposed by Ballé [22], which handles the non-differentiability of quantization by replacing it with a differentiable approximation. This is achieved by adding uniform noise during training and employing a learned prior for entropy modeling. Such techniques enable end-to-end optimization based on the rate-distortion objective, allowing the trade-off between the bitrate and reconstruction quality to be controlled by adjusting the distortion weight. Typically, these methods require separate models or network weights to operate at different target bitrates (e.g., low, medium, and high).

Although Ballé-based methods offer a practical workaround for quantization non-differentiability, they have some limitations. They often require training and maintaining multiple models to support different bitrates, making them less efficient for adaptive compression tasks. By contrast, our proposed VQ-HoloNet introduces a built-in bitrate tuner that enables dynamic control over the compression ratio within a single unified network. This flexibility allows the model to adjust the bitrate on demand—without the need for retraining or uploading additional weights—making it particularly well-suited for adaptive compression scenarios, such as QoS control over networks. In addition, [20] used a multi-network architecture, employing two U-Nets—one as a phase initializer and the other as a phase generator. This design increases the computational burden on the edge hardware, because an additional phase generator is required after decompression. However, our method reduces the computational load by integrating the necessary functionality into a more efficient single-network design.

In the following, we compare our proposed method with the state-of-the-art method of DPRC [16]. While [18] and [20] report relevant results, they are not considered for the comparison as the former utilizes a non-public dataset and the latter employs a distinct filed of view (FOV) centered on pupil gaze, which differs from our experimental objective. We set parameters and configuration as following. The reconstruction loss weights to $w_{\text{MSE}} = 1$, $w_{\text{MS-SSIM}} = 0.1$, and $w_{\text{WFFT}} = 0.025$ to balance the pixel-wise accuracy, perceptual similarity, and frequency-domain fidelity.

For both VQ-HoloNet (low) and VQ-HoloNet (ultra low) scenarios, we have similar following network configuration and parameters. The latent space is represented by a codebook. This codebook comprises $k = 4096$ code vectors, each represented by a 128-dimensional vector. The Seq2Seq model is trained to operate over a continuous range of code word counts, allowing $\tilde{K}$ to be any value between 512 and 4096 to allow compression at different bit rates. During inference, the framework can be any value between these two numbers. However, to align with standard hardware memory alignment, we limit the model to generating codebooks with sizes in powers of two ($\tilde{K} \in \{512, 1024, 2048, 4096\}$).

In our implementation, the EMA decay for codebook updates is set to 0.95 and the commitment cost is fixed at 0.25. Furthermore, for $N = 100$, the network is optimized using the Adam optimizer with a constant learning rate of $\eta = 1 \times 10^{-4}$. The original image size is $(H \times W) = 1072 \times 1920$. To reduce ringing artifacts after ASM propagation, we set the Region of Interest (RoI) to $700 \times 1400$ and apply an additional zero padding for stability, following the approach in [20]. In VQ-HoloNet (low) the latent space is $134 \times 240$ for the bottom level and $67 \times 120$ for the top level, while for VQ-HoloNet (ultra low) the latent spaces are $268 \times 480$ for the bottom level and $134 \times 240$ for the top level.

## 5.4 Quantitative Results

Table 2 compares the performance of three models—VQ-HoloNet (low), VQ-HoloNet (ultra low), and DPRC [16]—in terms of reconstruction quality and compression efficiency on the dataset Set1. Both VQ-HoloNet (low) and VQ-HoloNet (ultra low) achieve significantly lower Bpp values, while the perceptual quality is comparable or slightly reduced. VQ-HoloNet (low) achieves an average PSNR of 29.43 dB at a Bpp of 1.787, while DPRC (High quality) requires 3.5 Bpp (approximately 2 times compare to our low model) to achieve a similar quality. The VQ-HoloNet (ultra low) further reduces the Bpp by approximately 78% on average compared to the low model, while only decreasing PSNR by approximately 13% on average. As shown in Fig. 5, VQ-HoloNet (low) preserves the quality despite the use of fewer Bpp. These results demonstrate the effectiveness of the VQ-HoloNet models, which main-

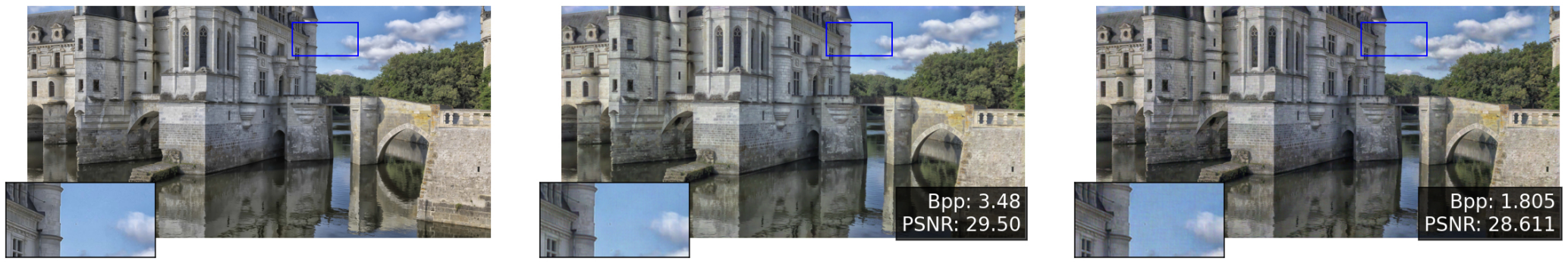

(a) Original amplitude image (b) DPRC (High Quality) (c) VQ-HoloNet (low)

Figure 5: A quantitative result comparing VQ-HoloNet with DPRC.

tain high visual fidelity even at substantially reduced bitrates. Compared to DPRC (High quality), VQ-HoloNet (low) reaches almost the same PSNR, but with nearly 50% lower Bpp with only a marginal 3% decrease in SSIM and 0.05 increase in LPIPS. In other words, while maintaining nearly equivalent PSNR and perceptual fidelity metrics, our method operates at half the rate of competing models in a shorter inference time.

The reported inference time encompasses both network decoding and entropy decoding for all the evaluated models on test Set1 across all channels.

Table 2: Quantitative comparison of the average reconstruction quality, compression efficiency, and average decoding time across three channels on Set1.

| **Method** | ↑ **PSNR** | ↑ **SSIM** | ↓ **LPIPS** | ↓ **Bpp** | ↓ **Time** |
|---|---|---|---|---|---|
| VQ-HoloNet (low) | 29.43 | 0.84 | 0.25 | 1.79 | 0.03s |
| VQ-HoloNet (Ultera low) | 26.00 | 0.7 | 0.4 | 0.4 | 0.05s |
| DPRC (High Quality) | 29.67 | 0.87 | 0.2 | 3.50 | 0.116s |
| DPRC (Mid Quality) | 28.61 | 0.85 | 0.22 | 2.11 | 0.116s |
| DPRC (low Quality) | 27.01 | 0.78 | 0.301 | 1.05 | 0.115s |

Table 3: Quantitative comparison of the reconstruction quality and compression efficiency across three channels on Set2.

| Method | ↑ PSNR | ↑ SSIM | ↓ LPIPS | ↓ Bpp | ↓ Time |
|---|---|---|---|---|---|
| VQ-HoloNet (low) | 32.71 | 0.88 | 0.23 | 1.70 | 0.03s |
| DPRC (High Quality) | 33.39 | 0.91 | 0.18 | 3.02 | 0.116s |
| DPRC (Mid Quality) | 31.34 | 0.89 | 0.20 | 1.75 | 0.114s |
| DPRC (low Quality) | 29.74 | 0.85 | 0.30 | 0.85 | 0.113s |

While the DPRC network exhibits increased decode latency owing to the hyperprior model computations for its latent space, VQ-HoloNet avoids such an overhead. VQ-HoloNet uses a lookup table to map indices to the data, making retrieving and decoding extremely fast. As reported in Table 2, VQ-HoloNet (ultra low) uses a deeper architecture than the (low) variant; this results in a marginal 0.02s increase in decoding time in exchange for a significantly lower Bpp trade-off.

As reported in Table 3, compared to DPRC (Mid quality), our approach attains a higher PSNR while maintaining comparable bitrates, whereas in comparison with DPRC (High quality), our method achieves a lower bitrate with a comparable PSNR. Fig. 6 demonstrates the superiority of our method over DPRC on the Set2 dataset, in achieving a higher quality at lower bitrates. With VQ-HoloNet (low), our method achieves significantly lower Bpp values while maintaining quality close to that of DPRC (High Quality) model, which relies on a larger latent space and higher quantization for optimal fidelity.

Our VQ-HoloNet does not utilize the entire 4096 code vectors uniformly. Fig. 7 illustrates the frequency of each code vector index averaged on the test Set1 for each channel. As shown, both the top and bottom layers of each channel exhibit non-uniform code vector selection, where some code vectors occur far more frequently than others. This imbalance in the code vector distribution highlights the potential of entropy coding methods, such as ANS, to further improve compression efficiency. Specifically, during inference time, we utilize for ANS entropy decoding, which achieves near-optimal compression rates in a short time.

Fig. 8 presents the rate–distortion performance of our proposed method compared with several existing techniques, evaluated using PSNR, SSIM [36], and LPIPS [37] metrics on Set1. The results are plotted against Bpp to illustrate the trade-off between the compression rate and visual quality. Our approach is tested under two compression modes: VQ-HoloNet (low) and VQ-HoloNet (ultra low). Even at very low bitrates, our method maintains competitive or superior visual quality, as indicated by the higher PSNR and SSIM values and lower LPIPS scores. According to Fig. 8, compared to conventional codecs combined with iterative or neural CGH methods—specifically SGD + JPEG, WirtingerSGD+JPEG, and HoloNet + JPEG- our framework demonstrates a superior rate–distortion balance. Both SGD-based methods undergo 500 iterations to ensure the convergence.
The curves indicate that our framework is highly competitive with DPRC, the current state-of-the-art. Specifically, our method (low) achieves an enhancement of −33.91% BD-Rate and BD-PSNR [38] of 1.02 dB.

In this comparison, WirtingerSGD demonstrates superiority over SGD in terms of PSNR and SSIM metrics while maintaining slightly lower perceptual scores on LPIP. Although previous work by [16, 9] indicates that during hundreds of iterations, SGD overcomes neural network based meth-

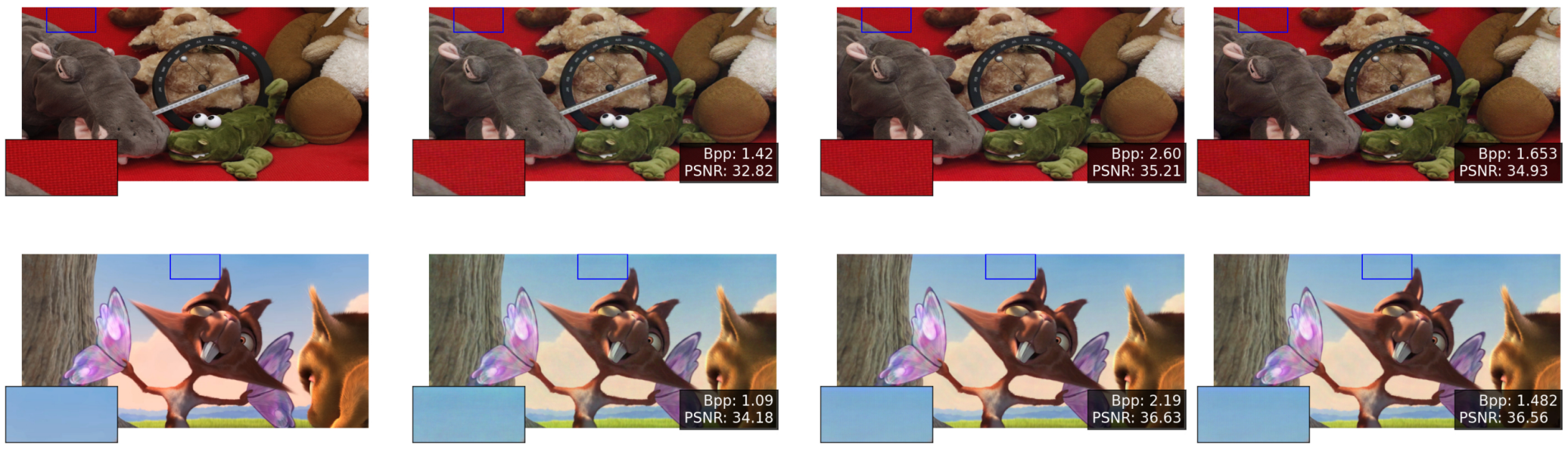

(e) Original image (f) DPRC (Mid Quality) (g) DPRC (High Quality) (h) VQ-HoloNet (low)

Figure 6: A quantitative result comparing VQ-HoloNet with DPRC.

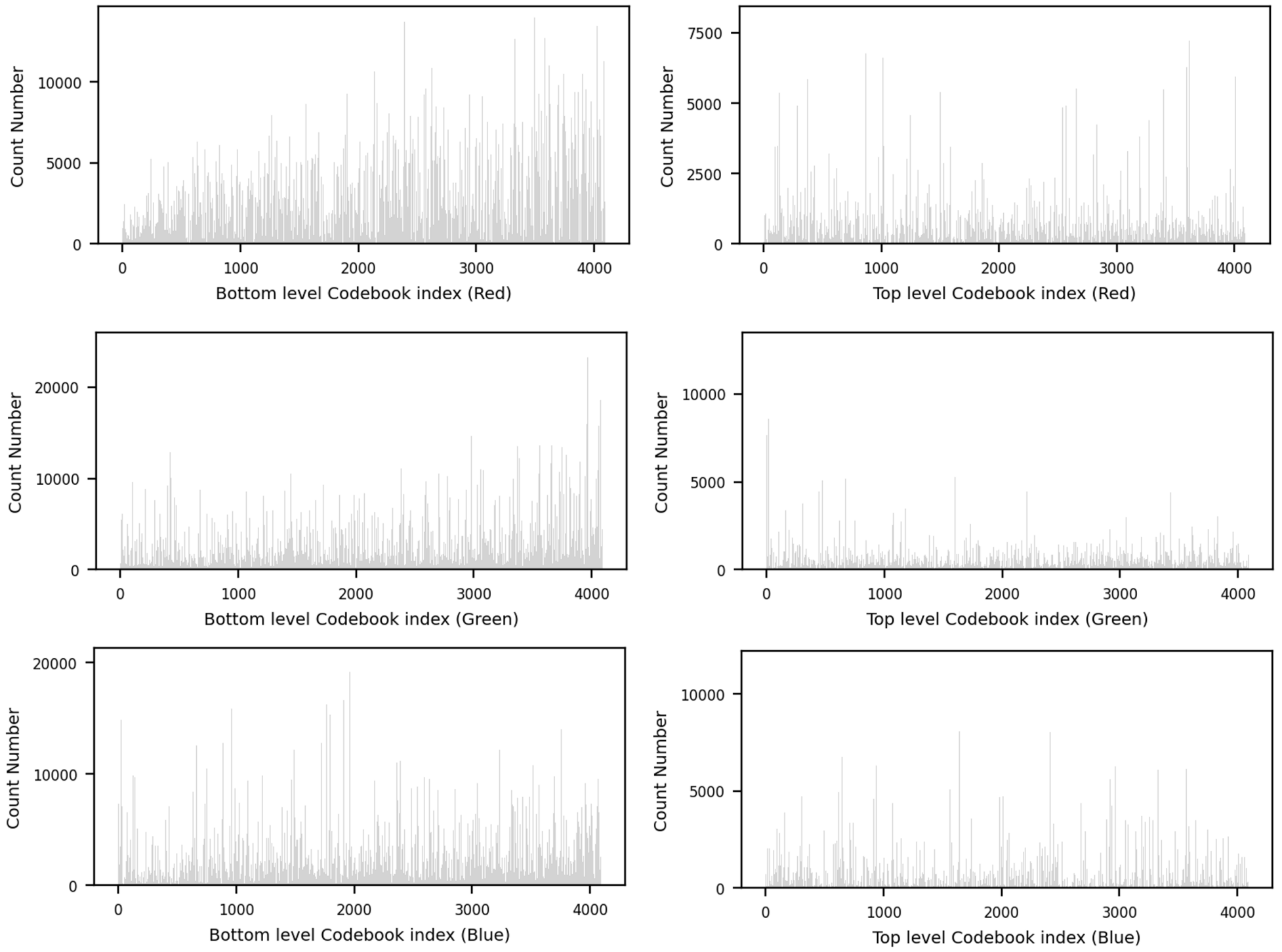

Figure 7: For codebook size of 4096, the codebook utilization, or the frequency of indices selected in each layer of the bottom and top level for each channel across the test Set1.

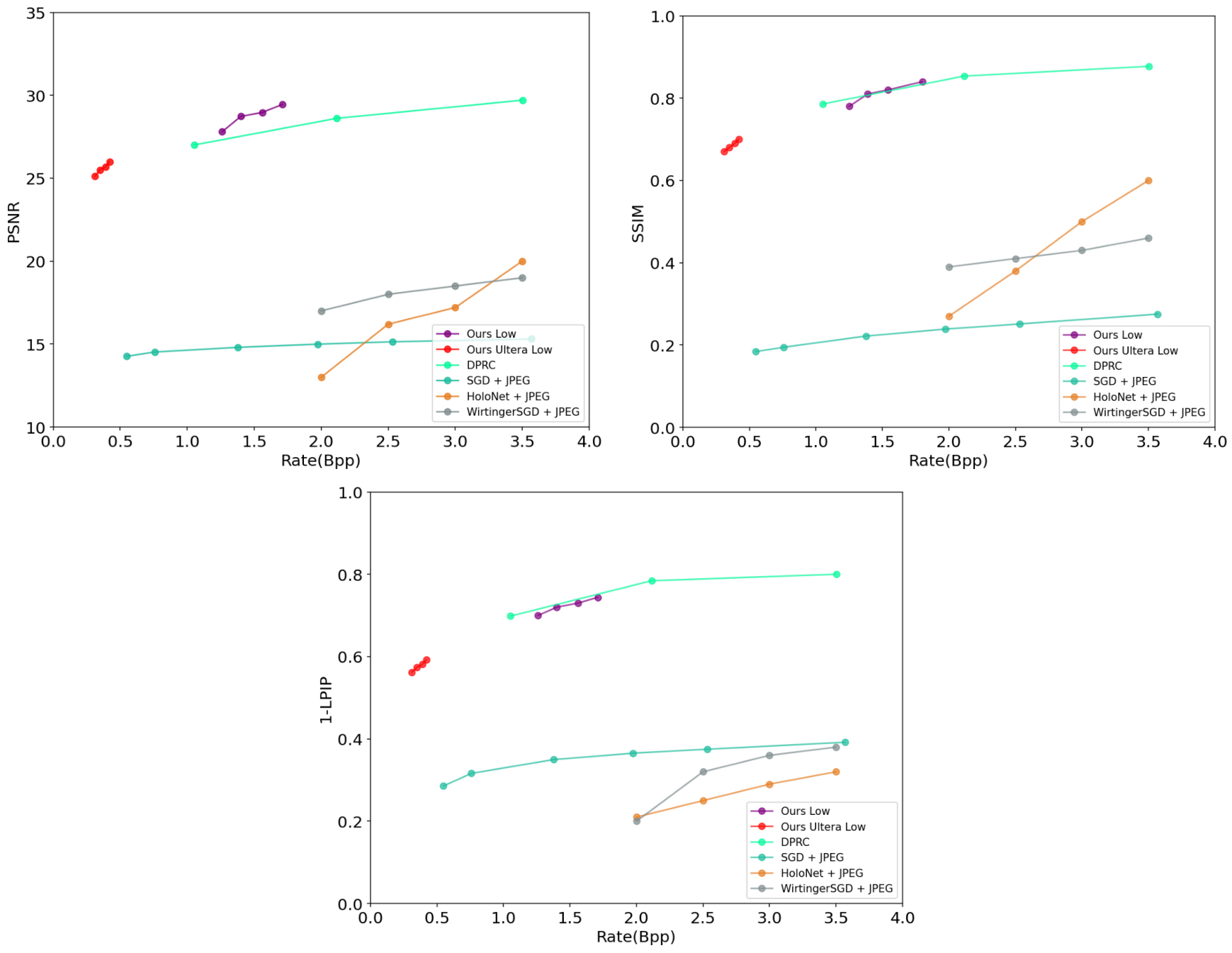

Figure 8: On Set1 dataset, rate-distortion performance indicated as quality metrics vs. rate (Bpp)

ods for phase-only hologram generation, its integration with JPEG compression reveals a significant incompatibility. Specifically, the PSNR for these JPEG-compressed methods remains below 20 dB, and the SSIM fails to exceed 0.5. This performance collapse suggests that the structured artifacts introduced by DCT-based quantization in JPEG have particularly negative impact to the high frequency phase-only pattern produced by these iterative methods. While HoloNet coupled with JPEG shows a trend with upward slope at higher bitrates, its performance remains substantially inferior to our proposed method across the entire Bpp range.
Consequently, our method is well-suited for holographic image compression in bandwidth-limited scenarios, as it facilitates high-ratio compression without compromising the perceptual or numerical quality of the reconstructed object.

## 5.5 Qualitative Result

Fig. 9 presents a qualitative and quantitative comparison of the four image reconstruction approaches: (a) Original Amplitude, (b) SGD + JPEG, (c) DPRC, and (d) Our approach. In the first row, our method (last column) successfully recovers the fine fur details of the lion, which are noticeably lost in the SGD+JPEG method or have a lower contrast in the DPRC method. It also achieves the highest PSNR (35.23 dB) with a competitive bit rate (1.729 Bpp). In the second row, our method (last column) successfully recovers fine facial details—such as wrinkles between the eyebrows with enhanced contrast, which are noticeably blurred or lost in other methods. It also achieves the highest PSNR (34.281 dB) with a competitive bit rate (1.773 Bpp).

In the third row, the parrot image reconstructed by our method preserves vibrant colors and textures, achieving a PSNR (34.400 dB) similar to that of DPRC while requiring nearly 1 Bpp less (1.630 vs. 2.620).

Finally, the fourth row demonstrates the effectiveness of the proposed method in reconstructing dark regions with high perceptual quality, achieving nearly the same PSNR (32.654 dB) as DPRC, while using significantly fewer bits (1.740 Bpp vs. 3.120 Bpp). Overall, these results highlight the strength of our approach in preserving visual detail and quality across diverse scenes, even under tight bitrate constraints.
We also evaluated the performance of our VQ-HoloNet (low) and VQ-HoloNet (ultra low) compression schemes at different bit rates. The results are shown

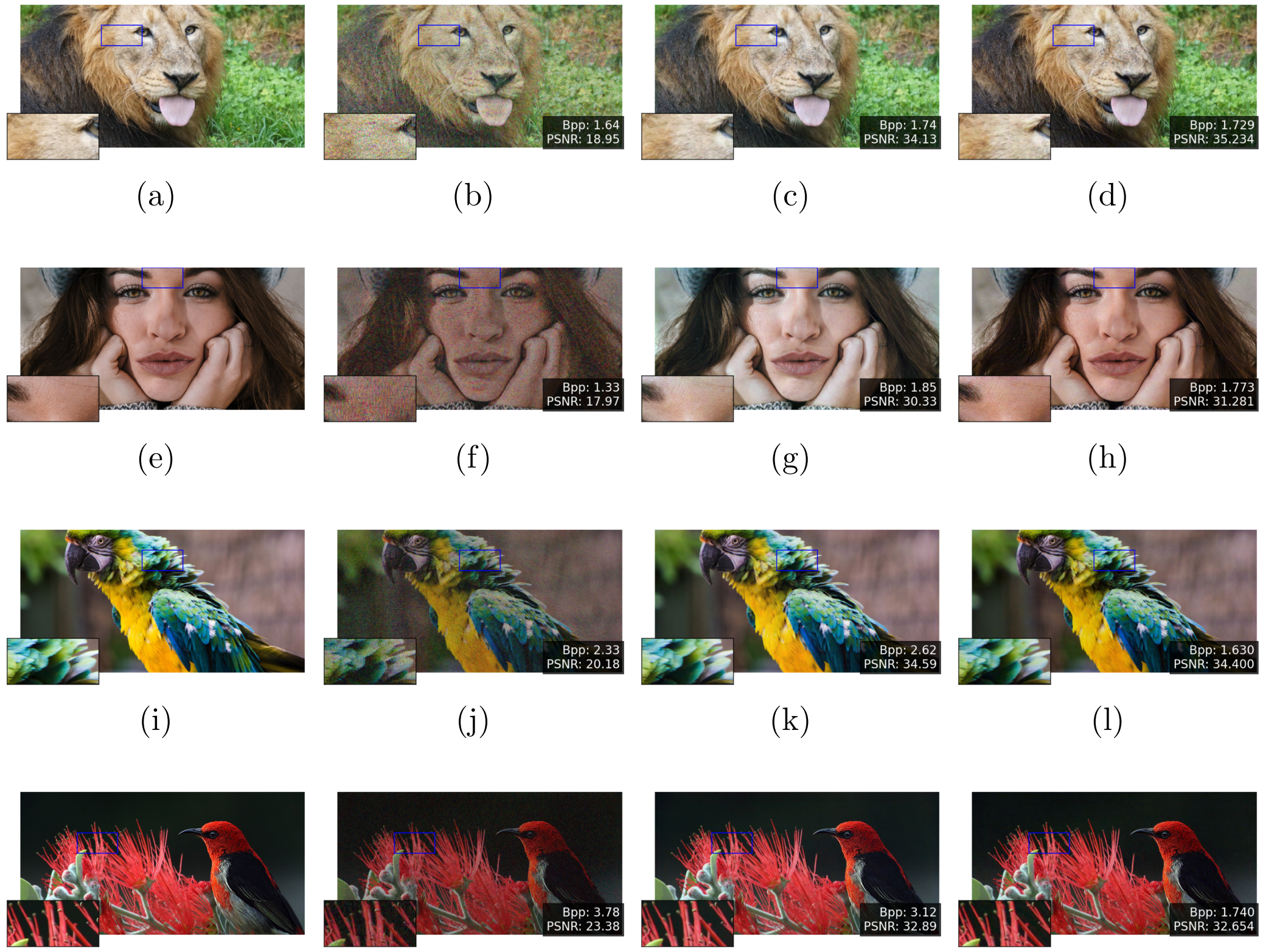

(a) (b) (c) (d)

(e) (f) (g) (h)

(i) (j) (k) (l)

(m) Original Amplitude (n) SGD + JPEG (o) DPRC (mid, high) (p) Ours

Figure 9: Numerical reconstruction. Comparison between three compared methods on images from [35]. The first two rows show our high-quality results versus DPRC's medium-quality outputs. The last two rows compare our method with DPRC's high-quality outputs. The baseline method (SGD + JPEG) exhibits a high amount of speckle noise.

for two images of a butterfly and a statue in Fig. 10. As can be seen from the figures, in both modes (VQ-HoloNet (low) and VQ-HoloNet (ultra low)) the quality of the decoded and reconstructed image improves by increasing the codebook size (i.e. the bit rate). The quality improvement is evident in sharper edges in the zoom-in panes.

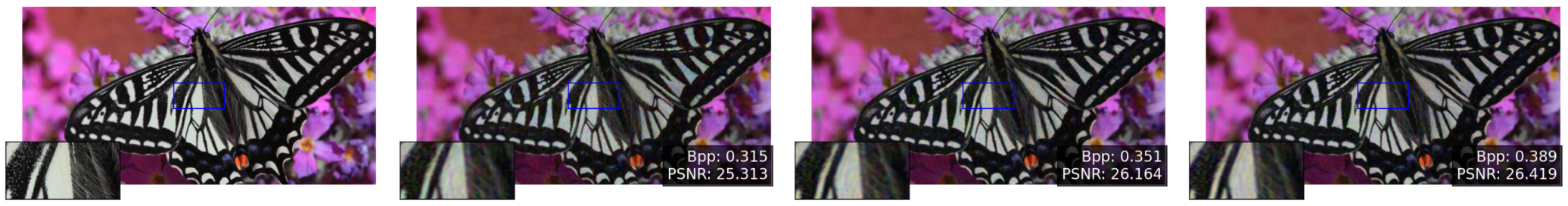

(a) Original amplitude (b) ultra low, codebook size = 512 (c) ultra low, codebook size = 1024 (d) ultra low, codebook size = 2048

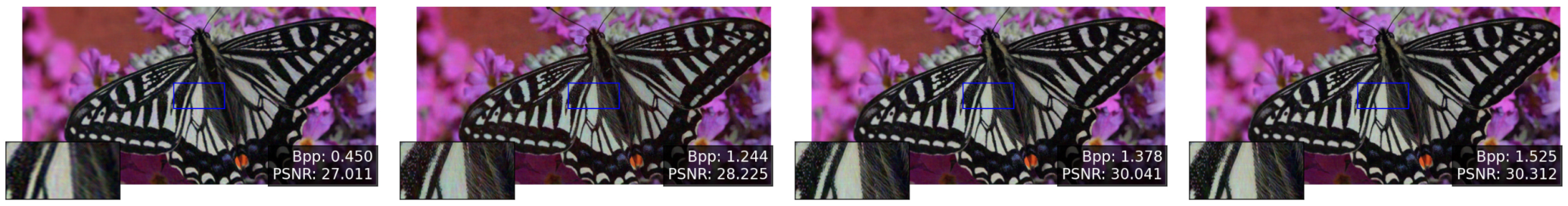

(e) low, codebook size = 512 (f) low, codebook size = 1024 (g) low, codebook size = 2048 (h) low, codebook size = 4096

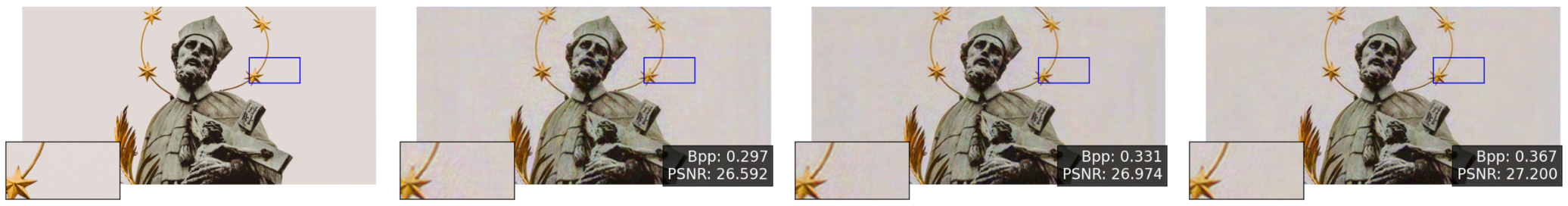

(i) Original amplitude (j) ultra low, codebook size = 512 (k) ultra low, codebook size = 1024 (l) ultra low, codebook size = 2048

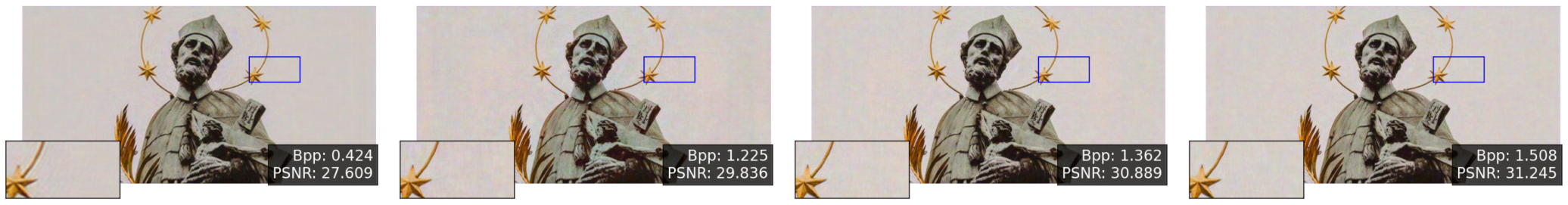

(m) low, codebook size = 512 (n) low, codebook size = 1024 (o) low, codebook size = 2048 (p) low, codebook size = 4096

Figure 10: Numerical reconstruction. RAVQ-HoloNet at multiple Bpp levels, showing low and ultra low networks applied to the butterfly and statue samples from [35].

# 6 Hardware Experimental Results

In the preceding sections, we compared software-based results and reported their respective metrics. Here, we present reconstructions from RAVQ-HoloNet as displayed through an SLM. These are compared against results from DPRC and SGD+JPEG at a comparable bitrate of approximately 0.5 bpp, as illustrated in Fig. 11. Across all test images, our method demonstrates superior contrast and sharper object edges. While DPRC produces smoother background regions, it lacks the high-frequency detail of our approach. As evidenced in the penguin example (last row), the primary limitation of RAVQ-HoloNet lies in smooth, bright region such as the penguin's body where some degradation occurs. Conversely, the SGD method suffers from severe reconstruction artifacts due to the heavy JPEG compression required to meet such a low bitrate.

The optical setup used for the measurements is illustrated in Fig. 12. A high-power M2 Titanium Sapphire laser with a wavelength of 813 nm is used for holography. The output beam of the laser is highly elliptical and astigmatic. Using a beam-shaping module comprised of cylindrical lenses, the beam is transformed into a 6 mm circular Gaussian beam [39]. The beam is then expanded using a 2:3 telescope (lenses L1 and L2) to produce a 9 mm Gaussian beam. A pair of $\lambda/2$ waveplates together with a polarization beam splitter (PBS1) is used to purify the laser polarization and align it with the liquid-crystal director axis of the SLM (model Meadowlark, P1920).

The beam is then incident on the SLM with the test phase uploaded, and the hologram is formed 13.5 mm away the SLM. The beam subsequently passes through a $4f$ system (lenses L3 and L4), which relays the SLM plane while enabling Fourier filtering of the zero-order beam in the focal plane between L3 and L4. The output beam is then polarization filtered using another polarization beam splitter (PBS2) to remove residual zero-order light. Finally, the hologram plane is relayed by a 2:1 telescope (lenses L5 and L6) and recorded by a CMOS camera (Basler, a2A2840-48umBAS).

The recorded images exhibit Fabry–Perot fringes that degrade the image quality. These fringes arise from the high coherence of the laser source and reflections between the camera sensor and the protective glass of the CMOS camera. As a result, the measured intensity is effectively multiplied by a spatial modulation that introduces periodic bright and dark fringes in the captured image.

To correct for this effect, a uniform phase is uploaded to the SLM so that the beam appears Gaussian at the camera plane. The recorded image therefore corresponds to a Gaussian beam modulated by periodic fringes. The underlying Gaussian profile is estimated by applying Gaussian blurring followed by an ellipse fit to the contour of the resulting image. The multiplicative modulation function is then obtained by dividing the recorded image by the estimated Gaussian profile. This factor is subsequently used to compensate for the Fabry–Perot fringes present in the hologram measurements.

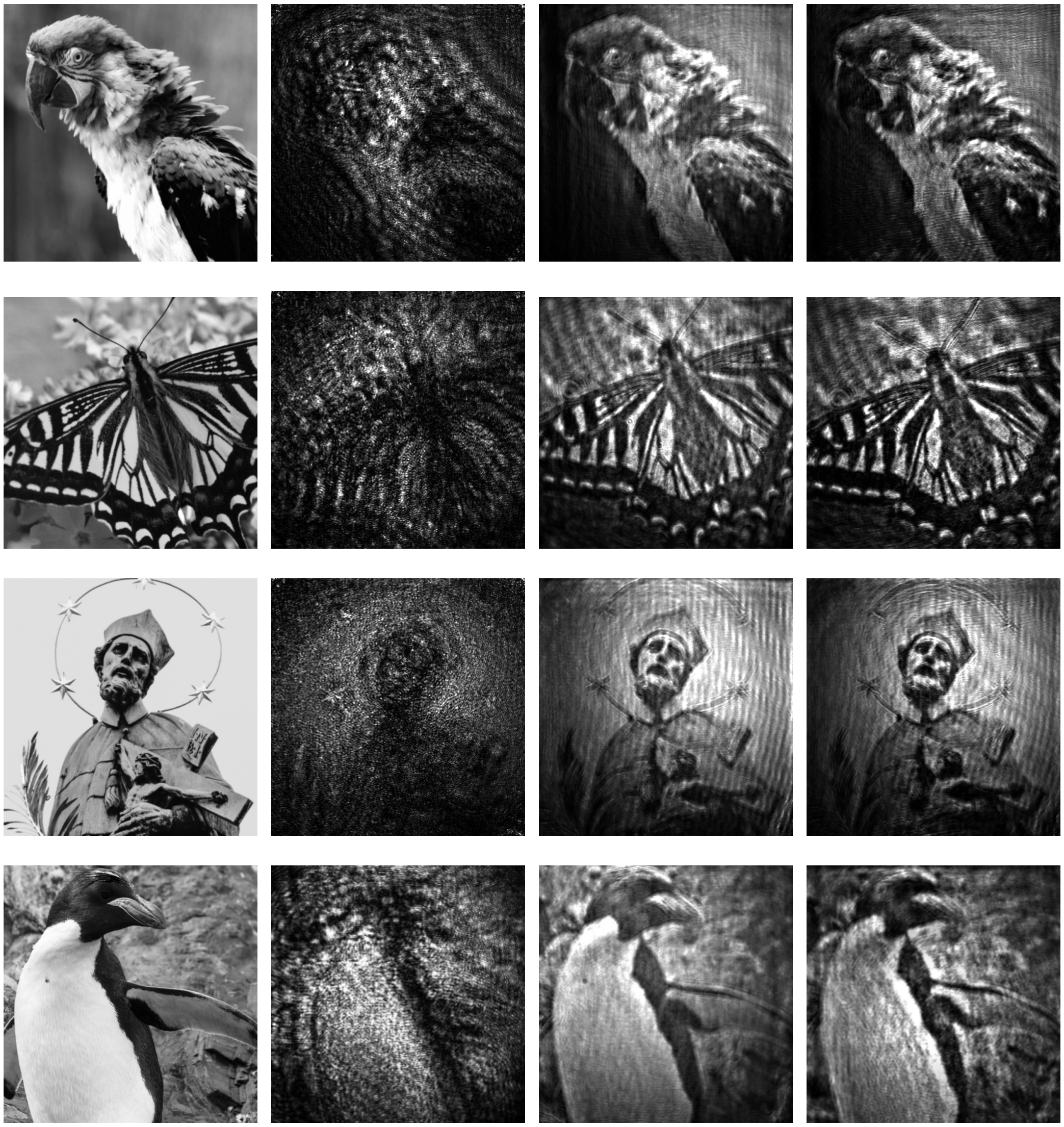

Figure 11: Experimental results using SLM. Columns display the original image, SGD+JPEG, DPRC, and our results respectively from left to right.

To maintain parity between the digital simulation and the physical implementation, the RoI and pixel pitch were kept identical to the software model. The only primary deviations in the hardware setup involve the operating wavelength and propagation distance, necessitated by the available laboratory laser source and the mechanical constraints of the optical bench.

Specifically, while the logical pixel pitch remains consistent, the input images were digitally scaled to match the physical active area of the SLM. Furthermore, the propagation distance was fixed according to the laser's pre-aligned position due to limitations in adjusting the optical path length. Finally, as the camera's FOV was smaller than the full reconstruction area, the captured region was restricted to a specific subset of the reconstructed object, as illustrated in Fig. 11.

# 7 Future work

Future extensions of this work may incorporate salient [40] or gaze-driven regions [20, 41] to guide the reconstruction process. This would enable the network to allocate more code vectors to perceptually important areas while tolerating greater artifacts in non-salient regions, thereby improving both the perceptual quality and the compression efficiency. In addition, because RGB channels share similar structural components, jointly compressing complex-valued RGB holograms is expected to be more efficient than training separate networks for each channel. In this study, we applied the proposed compression method to 2D holography. To extend this toward 3D holography, modalities such as RGB-D, 2.5D, and full 3D representations can be explored [42]. For example, the generator proposed in [5] can be integrated into our framework, as it is capable of producing complex-valued holograms from RGB-D inputs with continuous depth. Furthermore, approaches such as holographic stereograms or holography without Fresnel propagation (e.g.,

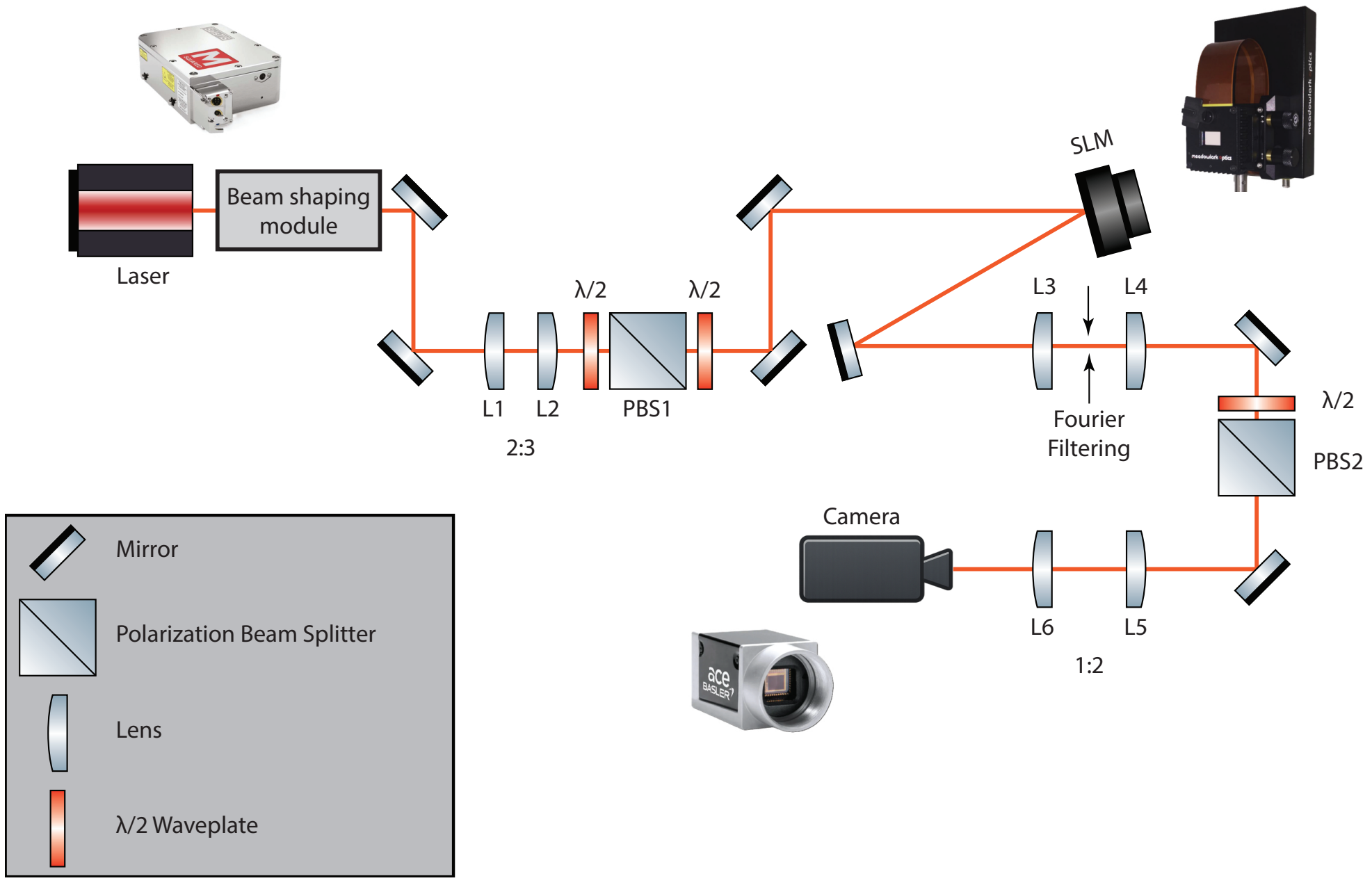

Figure 12: Hardware setup for hologram display.

[43]) employ different data formats, but can still be adapted to our framework as a combination of multiple 2D views. When working with RGB-D inputs, depth-aware propagation using models, such as [2] or ASM applied to masked depth maps, may yield more realistic reconstructions.
Moreover, techniques such as camera-in-the-loop holography [9] and calibration-aware pipelines can further help bridge the gap between simulated and real-world holographic displays.
Moreover, for industrial physics research, there is growing interest in developing new methods for manipulating laser beams in optical tweezer platforms used in atomic physics for trapping and transporting individual neutral atoms. Such systems enable precise control of atomic arrays and have important applications in quantum simulation [44] and quantum sensing [45]. In these platforms, holographic beam shaping can be used to generate and dynamically reconfigure arrays of optical traps. However, due to the computational overhead associated with hologram generation and the relatively slow refresh rates of conventional SLMs, dynamic trap manipulation is often implemented using acousto-optic deflectors (AODs) [46, 47, 48, 49]. While AOD-based systems provide fast beam steering, they are limited in their ability to generate arbitrary spatial light distributions. Recent advances in high-speed SLM technology have begun to bridge this gap, enabling rapid updates of holographic phase patterns and increasing the feasibility of fully holographic trap control [50]. These developments create a growing demand for faster and more efficient algorithms for hologram construction and compact representations of phase patterns. In particular, applications that require real-time holographic updates motivate the development of methods capable of efficiently encoding and transmitting holographic information under bandwidth and latency constraints.

# 8 Conclusion

In this study, we presented RAVQ-HoloNet, an unsupervised, rate-adaptive framework for compressing and reconstructing complex-valued holographic data. Our approach unifies complex-valued encoding and a phase-generation decoder within a single network, enabling end-to-end learning from raw hologram inputs to phase-only reconstructions. By integrating a Seq2Seq-based rate-adaptive vector quantization module into the VQ-HoloNet architecture, our method overcomes the limitations of fixed-rate compression, allowing multiple bitrates to be supported within a single trained model. The experimental results demonstrate that RAVQ-HoloNet achieves a competitive or superior reconstruction quality compared to existing state-of-the-art methods, while operating at significantly lower Bpp rates. Both quantitative metrics (PSNR, SSIM, and LPIPS) and qualitative visual comparisons confirm our model ability to preserve fine details and perceptual fidelity, even under aggressive compression settings. The flexibility of our rate-adaptive design eliminates the need to store and maintain separate encoder–decoder pairs for each bitrate, making it highly suitable for deployment on bandwidth- and storage-constrained devices. Beyond 2D holography, our framework can be extended to 3D modalities such as RGB-D, 2.5D, and full 3D holographic representations, as well as integrated with gaze-driven or saliency-aware compression strategies. These directions, along with real-world calibration and camera-in-the-loop adaptations, represent promising avenues for future research.

# Declaration

The authors used ChatGPT (OpenAI) to assist with language editing and sentence refinement during the writing process. All content was reviewed and edited by the authors.

# References

[1] A. Gilles, Fast hologram synthesis methods for realistic 3D visualization, Theses, INSA de Rennes (Sep. 2016).
URL `https://theses.hal.science/tel-01392677`

[2] S. Choi, M. Gopakumar, Y. Peng, J. Kim, G. Wetzstein, Neural 3d holography: learning accurate wave propagation models for 3d holographic virtual and augmented reality displays, ACM Transactions on Graphics 40 (6) (2021) 1–12.

[3] K. Matsushima, T. Shimobaba, Band-limited angular spectrum method for numerical simulation of free-space propagation in far and near fields, Optic Express 17 (22) (2009) 19662–19673.

[4] K. Matsushima, S. Nakahara, Extremely high-definition full-parallax computer-generated hologram created by the polygon-based method, Applied Optics 48 (34) (2009) H54–H63.

[5] L. Shi, B. Li, C. Kim, P. Kellnhofer, W. Matusik, Towards real-time photorealistic 3d holography with deep neural networks, Nature 591 (7849) (2021) 234–239.

[6] L. Shi, B. Li, W. Matusik, End-to-end learning of 3d phase-only holograms for holographic display, Light: Science & Applications 11 (1) (2022) 247.

[7] C.-K. Hsueh, A. A. Sawchuk, Computer-generated double-phase holograms, Applied Optics 17 (24) (1978) 3874–3883.

[8] A. Maimone, A. Georgiou, J. S. Kollin, Holographic near-eye displays for virtual and augmented reality, ACM Transactions on Graphics 36 (4) (2017) 1–16.

[9] Y. Peng, S. Choi, N. Padmanaban, G. Wetzstein, Neural holography with camera-in-the-loop training, ACM Transactions on Graphics 39 (6) (2020) 1–14.

[10] M. H. Eybposh, et al., Deepcgh: 3d computer-generated holography using deep learning, Optic Express 28 (18) (2020) 26636–26650. doi:10.1364/OE.399624.
URL `https://opg.optica.org/oe/abstract.cfm?URI=oe-28-18-26636`

[11] R. Horisaki, R. Takagi, J. Tanida, Deep-learning-generated holography, Applied Optics 57 (14) (2018) 3859–3863.

[12] R. Horisaki, Y. Nishizaki, K. Kitaguchi, M. Saito, J. Tanida, Three-dimensional deeply generated holography, Applied Optics 60 (4) (2021) A323–A328.

[13] R. W. Gerchberg, W. O. Saxton, A practical algorithm for the determination of phase from image and diffraction plane pictures, Optik 35 (2) (1972) 237–246.

[14] P. Chakravarthula, Y. Peng, J. Kollin, H. Fuchs, F. Heide, Wirtinger holography for near-eye displays, ACM Transactions on Graphics 38 (6) (2019) 1–13.

[15] B. Tobias, et al., Jpeg pleno holography presents the numerical reconstruction software for holograms: an excursion in holographic views, Applied Optics 62 (10) (2023) 2462–2469.

[16] Y. Wang, P. Chakravarthula, Q. Sun, B. Chen, Joint neural phase retrieval and compression for energy-and computation-efficient holography on the edge, ACM Transactions on Graphics 41 (4) (2022).

[17] M. Zhou, et al., End-to-end compression-aware computer-generated holography, Optic Express 31 (26) (2023) 43908–43919. doi:10.1364/OE.505447.
URL `https://opg.optica.org/oe/abstract.cfm?URI=oe-31-26-43908`

[18] H. Ko, H. Y. Kim, Deep learning-based compression for phase-only hologram, IEEE Access 9 (2021) 79735–79751.

[19] L. Shi, R. Webb, L. Xiao, C. Kim, C. Jang, Neural compression for hologram images and videos, Opt. Lett. 47 (22) (2022) 6013–6016. doi:10.1364/OL.472962.
URL https://opg.optica.org/ol/abstract.cfm?URI=ol-47-22-6013

[20] Z. Dong, Y. Ling, C. Xu, Y. Li, Y. Su, Gaze-contingent efficient hologram compression for foveated near-eye holographic displays, Displays 79 (2023) 102464.

[21] J. Ballé, V. Laparra, E. P. Simoncelli, End-to-end optimized image compression, arXiv preprint arXiv:1611.01704 (2016).

[22] J. Ballé, D. Minnen, S. Singh, S. J. Hwang, N. Johnston, Variational image compression with a scale hyperprior, arXiv preprint arXiv:1802.01436 (2018).

[23] J. Seo, J. Kang, Raq-vae: Rate-adaptive vector-quantized variational autoencoder, arXiv preprint arXiv:2405.14222 (2024).

[24] R. Ali, et al., Generating diverse high-fidelity images with vq-vae-2, Advances in Neural Information Processing Systems 32 (2019).

[25] K. He, X. Zhang, S. Ren, J. Sun, Deep residual learning for image recognition, in: Proceedings of the IEEE conference on computer vision and pattern recognition, 2016, pp. 770–778.

[26] J. Seo, J. Lee, J. Lee, H. Ko, Deep compression network for enhancing numerical reconstruction quality of full-complex holograms, Optic Express 31 (15) (2023) 24573–24597.

[27] V. D. Oord, et al., Neural discrete representation learning, Advances in Neural Information Processing Systems 30 (2017).

[28] S. Hochreiter, J. Schmidhuber, Long short-term memory, Neural Computation 9 (8) (1997) 1735–1780.

[29] I. Sutskever, O. Vinyals, Q. V. Le, Sequence to sequence learning with neural networks, Advances in Neural Information Processing Systems (NeurIPS) (2014).

[30] R. J. Williams, D. Zipser, Learning to predict by the methods of temporal differences, Cognitive Science 8 (1) (1984) 37–64.

[31] J. Duda, Asymmetric numeral systems: entropy coding combining speed of huffman coding with compression rate of arithmetic coding, arXiv: Information Theory (2013).
URL `https://api.semanticscholar.org/CorpusID:13409455`

[32] J. W. Goodman, Introduction to Fourier Optics, 3rd Edition, Roberts and Company Publishers, Greenwood Village, CO, 2005.

[33] Z. Wang, E. Simoncelli, A. Bovik, Multiscale structural similarity for image quality assessment, in: The Thrity-Seventh Asilomar Conference on Signals, Systems & Computers, 2003, Vol. 2, 2003, pp. 1398–1402 Vol.2. doi:10.1109/ACSSC.2003.1292216.

[34] S. Czolbe, O. Krause, I. Cox, C. Igel, A loss function for generative neural networks based on watson's perceptual model, Advances in Neural Information Processing Systems 33 (2020) 2051–2061.

[35] E. Agustsson, R. Timofte, Ntire 2017 challenge on single image super-resolution: Dataset and study, in: Proceedings of the IEEE Conference on Computer Vision and Pattern Recognition Workshops, 2017, pp. 126–135.

[36] Z. Wang, A. C. Bovik, H. R. Sheikh, E. P. Simoncelli, Image quality assessment: from error visibility to structural similarity, IEEE Transactions on Image Processing 13 (4) (2004) 600–612.

[37] R. Zhang, P. Isola, A. A. Efros, E. Shechtman, O. Wang, The unreasonable effectiveness of deep features as a perceptual metric, in: Proceedings of the IEEE conference on Computer Vision and Pattern Recognition, 2018, pp. 586–595.

[38] G. Bjontegaard, Calculation of average psnr differences between rd-curves, ITU-T SG16, Doc. VCEG-M33 (2001).

[39] S. Khoubyarian, A. Mashko, A. Cooper, Correcting astigmatism and ellipticity in gaussian beams using a cylindrical lens pair with tunable focal lengths, Optic Express 33 (22) (2025) 47292–47320. doi:10.1364/OE.570745.
URL `https://opg.optica.org/oe/abstract.cfm?URI=oe-33-22-47292`

[40] L. Wang, H. Lu, X. Ruan, M.-H. Yang, Deep networks for saliency detection via local estimation and global search, in: Proceedings of the IEEE conference on computer vision and pattern recognition, 2015, pp. 3183–3192.

[41] S. Rafiei, S. Shirani, Saliency-guided progressive hologram generation, in: Frontiers in Optics + Laser Science 2025 (FiO, LS), Optica Publishing Group, 2025, p. FV2B.6. doi:10.1364/FIO.2025.FV2B.6.
URL `https://opg.optica.org/abstract.cfm?URI=FiO-2025-FV2B.6`

[42] D. Kim, S.-W. Nam, S. Choi, J.-M. Seo, G. Wetzstein, Y. Jeong, Holographic parallax improves 3d perceptual realism, ACM Trans. Graph. 43 (4) (Jul. 2024). doi:10.1145/3658168.
URL `https://doi.org/10.1145/3658168`

[43] E. Dashdavaa, A. Khuderchuluun, H.-Y. Wu, Y.-T. Lim, C.-W. Shin, H. Kang, S.-H. Jeon, N. Kim, Efficient hogel-based hologram synthesis method for holographic stereogram printing, Applied Sciences 10 (22) (2020) 8088.

[44] H. Bernien, S. Schwartz, A. Keesling, H. Levine, A. Omran, H. Pichler, S. Choi, A. S. Zibrov, M. Endres, M. Greiner, V. Vuletić, M. D. Lukin, Probing many-body dynamics on a 51-atom quantum simulator, Science 357 (6347) (2017) 191–195.

[45] T. Roscilde, M. Kumari, A. Cooper, F. Mezzacapo, Exponential onset of scalable entanglement via twist-and-turn dynamics in xy models, arXiv preprint arXiv:2507.08206 (2025).

[46] A. H. Dadpour, T. Khayrullin, F. Afiouni, R. E. Sabeh, A. E. Mouawad, I. E. Hajj, A. Cooper, Low-latency control system for feedback experiments with optical-tweezer arrays, Physical Review Applied 24 (2025) 024062. doi:10.1103/svtr-1rmd.

[47] F. Afiouni, R. El Sabeh, N. Nishimura, I. El Hajj, A. E. Mouawad, A. Cooper, Efficient algorithms to solve atom reconfiguration problems. iii. the bird and batching algorithms and other parallel implementations on gpus, Physical Review Applied 112 (2025) 023109. doi:10.1103/2rds-k4n9.

[48] R. El Sabeh, J. Bohm, Z. Ding, S. Maaz, N. Nishimura, I. El Hajj, A. E. Mouawad, A. Cooper, Efficient algorithms to solve atom reconfiguration problems. ii. assignment-rerouting-ordering algorithm, Physical Review Applied 108 (2023) 023108. doi:10.1103/PhysRevA.108.023108.
URL `https://link.aps.org/doi/10.1103/PhysRevA.108.023108`

[49] A. Cooper, S. Maaz, A. E. Mouawad, N. Nishimura, Parameterized complexity of reconfiguration of atoms, Algorithmica 86 (10) (2024) 3284–3308. doi:10.1007/s00453-024-01262-z.
URL `https://doi.org/10.1007/s00453-024-01262-z`

[50] I. H. A. Knottnerus, Y. C. Tseng, A. Urech, R. J. C. Spreeuw, F. Schreck, Parallel assembly of neutral atom arrays with an slm using linear phase interpolation, SciPost Physics 19 (4) (2025) 118. doi:10.21468/SciPostPhys.19.4.118.